\documentclass[11pt,a4paper]{article}

\usepackage[utf8]{inputenc}
\usepackage[T1]{fontenc}
\usepackage{amsmath, amssymb}
\usepackage{graphicx}
\usepackage{booktabs}
\usepackage[margin=1in]{geometry}
\usepackage{setspace}
\usepackage{hyperref}
\usepackage{url}
\usepackage{float}
\usepackage{caption}
\usepackage{subcaption}
\usepackage[authoryear,round]{natbib}
\setcitestyle{notesep={; }}
\usepackage{tabularx}

\title{Self-Supervised Lexical Representation Learning for Fast, Large-Scale Phylogenetic Inference}
\author{Tim Wientzek}
\date{\today}

\begin{document}
\setstretch{1.5}

\maketitle

\begin{abstract}
Computational phylogenetics has become an essential tool in historical linguistics, yet its application at a global scale remains constrained by two factors: the labor-intensive manual annotation of cognacy judgments required for character-based methods and the substantial computational cost of inference on large datasets. This paper introduces a fully self-supervised contrastive learning framework that learns lexical representations directly from raw IPA-transcribed wordlists, without requiring cognacy annotations, alignments, or additional expert input. The model employs a dual contrastive objective: a word-level loss that organizes phonetically similar forms into a coherent space, and an auxiliary language-level loss that encourages the lexical space to reflect broader phonological properties of languages. From the resulting word representations, pairwise language distances are derived and used to infer a global phylogenetic tree of 3,399 language varieties. The inferred tree achieves a generalized quartet distance (GQD) to the Glottolog reference tree competitive with multiple baselines, while requiring only minutes of computation on a standard notebook GPU. Furthermore, the same representations capture diachronic concept stability: variance in pairwise distances across languages yields stability rankings that correlate significantly with established rankings. Ablation studies confirm that both the language-level objective and the use of phonetic feature vectors improved the inferred trees topology with regards to GQD. The framework thus provides a computationally efficient and fully automatic alternative for large-scale phylogenetic inference and offers a unified representation supporting downstream analyses at both the language and concept level.
\end{abstract}

\section{Introduction}
Computational phylogenetics has become an established tool in historical linguistics and is increasingly used to investigate questions such as language subgrouping and the timing of language divergence \citep{greenhill2023}. State-of-the-art methods are predominantly character-based and commonly rely on Bayesian estimation or maximum likelihood. These methods typically require manually established cognacy judgements or sound correspondences to be converted into binary character matrices. Creating these datasets is a time-consuming and labor-intensive process that requires deep expertise in the respective language families. This introduces potential biases, since experts may already hold prior assumptions about language relationships, and different experts may disagree on how the same evidence should be encoded. It also makes exact replication more difficult \citep{rama2019,wu2023,snee2026}.

The same problem limits large-scale inference in two ways. First, expert cognacy judgements or equivalent character encodings are not available for many languages. Second, the computational cost of phylogenetic inference becomes substantial once thousands of languages are considered. The first problem can be addressed through automated cognacy annotation or sound correspondence detection, and several such approaches have been shown to produce phylogenetic trees of reasonable quality \citep{rama2018,hauser2024}. However, these methods still require large amounts of computational resources when applied to large datasets. For example, the automated character-based inference of \cite{jager2026} on more than 3000 languages required several days of computation even on a high-performance cluster. This limits not only the initial inference of large trees, but also their repeated reconstruction when the underlying data change or new languages are added.

Distance-based methods provide an alternative approach to phylogenetic inference. Here, pairwise distances between lexical forms from two languages are calculated and averaged across shared concepts, resulting in a language-by-language distance matrix. Distances can for example be based on Levenshtein distance \citep[LD]{levenshtein1965,muller2009} or pointwise mutual information \citep[PMI]{jager2018}. A tree is then constructed from the resulting matrix, grouping languages according to their overall lexical similarity. In contrast to character-based methods, distance-based approaches do not explicitly model an evolutionary history, meaning that they cannot directly represent, for example, when a particular innovation occurred. They are, however, less computationally expensive and can provide useful approximations for exploratory analyses and hypothesis testing \citep[e.g.,][]{jager2015}. With curated collections such as ASJP and Lexibank, which cover thousands of languages, large-scale phylogenetic inference is therefore considerably more accessible than with manually constructed character matrices \citep{wichmann2025,list2022}. Improving the quality of automatically derived language distances is consequently of interest in its own right.

The emergence of large language models (LLMs) and related neural architectures has led to an increasing use of learned language representations for representing linguistic information in a continuous vector space. As such, there is a growing body of curated collections of language distances \citep{littell2017,khan2025,goot2025}, which have been used, for example, for predicting typological features \citep{amirzadeh2025}, dependency parsing \citep{ustun2022}, and predicting the suitability of languages for transfer learning \citep{lin2019,adilazuarda2024,rice2025,blaschke2025}. Other work has investigated learned language representations themselves for related tasks \citep{malaviya2017,oncevay2020,vastl2020,ostling2023}. More relevant to historical linguistics, several studies have shown that neural language representations contain genealogical and areal signals even when the underlying models were not trained specifically for historical-linguistic tasks \citep{rabinovich2017,rama2020}.
The question of whether lexical representations can instead be learned directly from multilingual lexical data in a self-supervised framework and subsequently used across multiple historical-linguistic tasks remains less explored. Once trained, the same representations could also serve as a basis for further analyses, since pairwise language distances and concept-level statistics can be calculated directly from the lexical representations. This is also a clear advantage over traditional distance-based methods, where pairwise language distances are derived directly from surface form similarity and individual form information is not retained. This limits these approaches to tree inference alone.

The work at hand investigates to which degree lexical representation learning on datasets designed specifically for applications in computational historical linguistics can recover historically meaningful structure sufficiently well to support both language-level and concept-level downstream tasks. The first task is phylogenetic inference, where the learned lexical representations are used to derive a pairwise language distance matrix and reconstruct a large-scale phylogenetic tree. The second task is concept stability, where variation in the learned lexical distances is used to derive a ranking of concepts according to their diachronic stability. These tasks test different aspects of the learned representation. Successful phylogenetic inference requires the representation to retain information about genealogical relationships between languages, while concept stability provides an independent test of whether the representation captures historically meaningful differences between lexical concepts.

\subsection{Related work}
\subsubsection{Automated large-scale tree inference}
Some of the earliest attempts at global-scale phylogenetic inference were based on the Automated Similarity Judgment Program \citep[ASJP]{wichmann2025}, a database containing wordlists for thousands of languages. The data include at least 40 core-vocabulary concepts for most languages, encoded in a strongly reduced set of sound classes. Trees were constructed using the neighbor-joining algorithm \citep{saitou1987} from pairwise averaged normalized Levenshtein distances \citep{muller2009} and have since been updated several times. The authors acknowledged that lexical similarity, which determines the grouping of languages in such a tree, does not necessarily reflect genealogical relatedness. Similarity may also result from lexical borrowing, universal tendencies such as onomatopoeia, or chance, especially for short word forms.

\citet{jager2018} subsequently used the ASJP database, which covered around 7000 languages at the time, to perform fully automated global-scale tree inference with three different approaches. Two of these were character-based, using automated cognacy judgments and per-concept binarized sound classes. The third was distance-based and used pointwise mutual information, a weighted and therefore more fine-grained measure of word similarity than Levenshtein distance. They could show that both the combination of both character-based methods and the distance-based method produced results that were similar across a range of evaluations and resulted in a tree close to the Glottolog expert tree \citep{glottolog2025}.

More recently, \citet{jager2026} applied a character-based approach to the Lexibank datasets \citep{list2022}. The study included languages with data for at least 40 of 210 predefined core vocabulary concepts, resulting in a sample of more than 3000 languages. The data were first converted from IPA to a variant of the ASJP script, significantly reducing the number of distinct characters. Multiple sequence alignments were then constructed for individual concepts and converted into a binary character matrix. The reduction in sound classes was necessary to keep maximum likelihood tree inference computationally feasible. Based on the resulting character matrix, Jäger obtained a tree topology that was highly similar to the Glottolog reference tree. Although not explicitly tested on the same dataset, they conclude that this method outperformed the previously proposed automated methods in \citep{jager2018} on the reported evaluations. 

The method presented by \citet{jager2026} therefore represents a strong large-scale baseline for automated phylogenetic inference. At the same time, it suffers severely from the high computational cost. Even with the reduced character representation and a further reduced concept set, tree inference required several days of computation on a high-performance cluster. Repeated inference therefore requires the utmost care and preparation.

\subsubsection{Genetic signal from neural encodings}
A separate line of research has investigated whether neural language representations contain genealogical information. \citet{rabinovich2017}, \citet{ostling2017}, and \citet{bjerva2019}, among others, identified genetic signals in representations produced by neural models trained for tasks such as machine translation and using dedicated language tokens. These results suggest that neural models can encode information about language relationships even when genealogical structure is not an explicit training target.

\citet{rama2020} investigated this question in greater detail using a multilingual BERT model. For each language represented in the model, translations of a set of diachronically stable concepts were passed through the model. Cosine distances between the resulting word representations were calculated for each concept and language pair and then averaged to obtain language-level distances. Clustering these distances produced groups that showed a solid correspondence with expert language-family classifications. The authors also found that variation in the distances associated with individual concepts correlated significantly with established concept stability rankings.

\subsection{Contribution}
In this work, I present DualCWE (Dual Contrastive Word Encoder), a hierarchical contrastive representation learning framework. The model is trained on the same filtered Lexibank data used by \citet{jager2026}. This makes it possible to evaluate the learned representation directly against a recent large-scale phylogenetic baseline under identical data conditions. The model learns lexical representations from multilingual word forms and uses an auxiliary language-level contrastive objective to encourage the lexical space to reflect broader properties of the languages.\newline
The three main contributions are:
\begin{enumerate}
    \item I introduce a fully self-supervised representation-learning framework for multilingual lexical data.
    \item I demonstrate that the derived language distances support large-scale phylogenetic inference with a lower generalised quartet distance (GQD) to the Glottolog reference tree than the corresponding result reported by \citet{jager2026}. Additionally, the complete model and tree inference run in only a few minutes on a regular notebook GPU.
    \item I show that the same lexical representations capture concept stability, providing evidence that the learned space preserves historically meaningful lexical structure beyond genealogical information alone.
\end{enumerate}

The resulting representation therefore provides a computationally efficient alternative for large-scale phylogenetic inference while also supporting downstream analyses at both the language and concept level. All code, data, training configurations, and model checkpoints are publicly available at XY.

\section{Data \& Methods}
\subsection{Data}
To ensure comparability with \citet{jager2026}, the same data are used here. The study drew on 185 Lexibank datasets \citep{list2022} and selected languages with coverage of at least 40 out of 210 pre-defined core vocabulary concepts. Concept stability was assessed following the ranking of \citet{dellert2018}. After applying these filters, the resulting sample comprises 3399 language varieties spanning 311 language families, 109 of which consist of three or more languages, and covering all macroareas. All word forms are available as IPA transcriptions.

\citet{jager2026} ultimately restricted the character matrix used for tree inference to the 105 highest-stability concepts rather than the full set of 210. This reduction was motivated by a phylogenetic difficulty analysis and represents the best trade-off between the amount of available data and the clarity of the phylogenetic signal for maximum likelihood inference. In the present study, no such reduction was applied. The computational constraints that motivate it in a maximum likelihood framework do not apply here, and it is assumed that a broader concept set is generally advantageous for the model as long as the concepts included belong to the more stable ones (see \S\ref{sec:ablation} for an ablation). Given that the concepts were selected based on this condition, the full set of 210 concepts is used in the work at hand.

For evaluation, the inferred tree is compared against the Glottolog expert tree \citep{glottolog2025}, pruned to the languages present in the sample. The evaluation procedure is described in \S\ref{sec:eval}. 
The filtered dataset and the pruned Glottolog tree have been made publicly available by the original author and are additionally included in the supplementary materials of this paper \citep{suppmat}.

\subsection{Methods}
\subsubsection{Model and Training}
\label{sec:model}
\begin{figure}[!ht]
\centering
\includegraphics[width=0.8\textwidth]{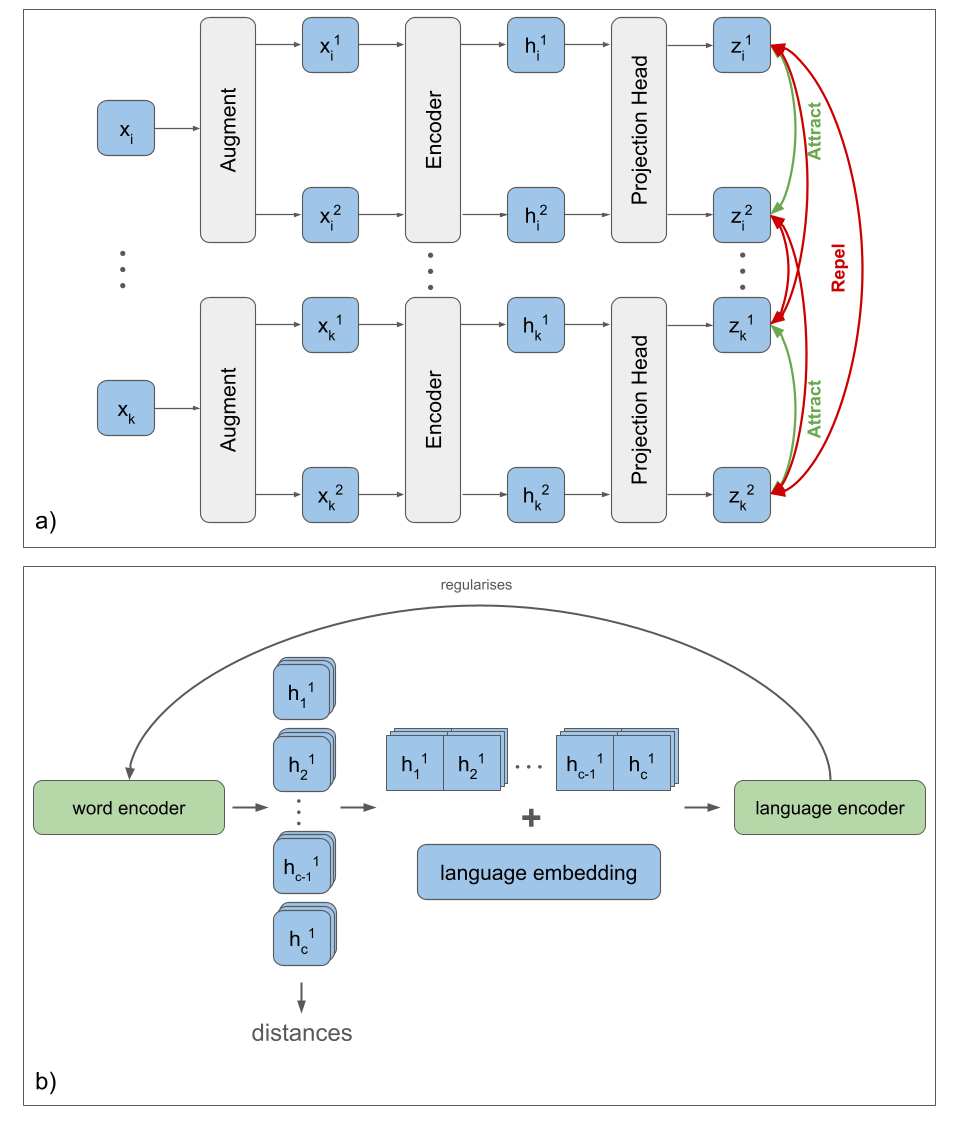}
\caption{Schematic model overview. a) shows the internal architecture of both the word and language encoder (adapted from \citet{wientzek2025b}). For the word encoder, each batch item $x$, a word, is passed through the model twice to get two augmented views, which are then processed by the encoder, pooled and passed through a projection head. The model objective (loss function) maximises agreement between the two projected augmented views of an item (attract), while minimising agreement between these and the projections from all other items (repel). For the language encoder, $x$ constitutes the concatenated lexical representations with the augmentations applied in the word encoder. b) shows the overall architecture. Batches of word forms are encoded into lexical representations and concatenated for the language encoder. Language embeddings are added to each position of the sequence. Both the language encoder and the evaluation operate on the outputs of the word encoder, before the projection layer.}
\label{fig:schematic}
\end{figure}

The model presented here builds on the contrastive learning framework introduced in \citet{wientzek2025b} for unsupervised borrowing detection, extending it with a second, language-level encoding component. Both the word encoder architecture and the contrastive loss function are taken from \citet{wientzek2025b} without modification and readers are referred to that work for a more detailed description of both. What follows describes the full architecture and training procedure, with emphasis on the extensions introduced here. A schematic overview of the model architecture is provided in \autoref{fig:schematic}.

Training batches are specifically constructed to contain a fixed number of $l$ languages with $c$ randomly sampled concepts each, yielding a batch of $l \times c$ word forms. This is necessary for the second part of the model, the language encoder (see below), which requires multiple word forms per language in a batch.

Input word forms are represented as sequences of phonetic feature vectors, with each IPA character encoded as a 39-dimensional vector capturing place and manner of articulation and related features. Feature vectors were generated using the Soundvectors library \citep{rubehn2024} for Python. The advantages of feature vectors over discrete characters has been demonstrated for phonological reconstruction \citep{wientzek2025a} and borrowing detection \citep{wientzek2025b} and testing suggests that it also holds for the work at hand (see \S\ref{sec:ablation}). These sequences are projected into a higher-dimensional embedding space via a linear transformation, followed by layer normalisation and input dropout. The same augmentations described in \citet{wientzek2025b} are then applied: characters may be duplicated with a fixed probability \citep[following][]{wu2022}, Gaussian noise may be added to the embeddings, and standard Transformer dropout is applied throughout training. A learned concept embedding - a vector associated with each concept in the dataset and updated during training - is then added to every position of the sequence before it is passed to the encoder. This provides the model with information about the semantic concept a word form refers to, therefore reducing the probability that cross-concept chance resemblances are pulled together during training. The augmented and concept-conditioned sequence is processed by a Transformer encoder \citep{vaswani2017} with relative positional encodings \citep{su2023}, and the final word-level representation is obtained by masked mean pooling over all non-padding positions. Following \citet{chen2020}, these encoder outputs are passed through a projection head for loss computation during training, while for evaluation the encoder outputs themselves are used.

The word representations produced by the word encoder are subsequently grouped by language. For each of the $l$ languages in the batch, the $c$ word representations are arranged into a sequence, transforming the batch from a flat set of $l \times c$ vectors into $l$ sequences of length $c$. A learned language embedding is added to each of the $c$ positions to identify the language to which the sequence belongs. This sequence is then processed by a second, identical Transformer encoder and aggregated by mean pooling over all positions. The resulting vector constitutes the language-level representation and is passed through a separate projection head for training.

Two views of each batch item are produced by passing the data through the model twice under different random augmentations. The model is trained using two separate contrastive losses, both taking the NT-Xent form described in \citet{wientzek2025b}. At the word level, the two projected representations of the same word form constitute a positive pair, with all other word forms in the batch serving as negatives. At the language level, the two projected representations of the same language - obtained by aggregating over the two augmented views of its word forms - constitute a positive pair, with all other languages in the batch as negatives. The total training loss is given by the sum of the word-level loss and the language-level loss weighted by a factor of 0.2.

The language encoder has a purely auxiliary function and its outputs are not used in any evaluations, which rely exclusively on the lexical representations. Its purpose is instead to shape the space in which word representations are organised. By training the model to produce consistent language-level representations across augmented views and keeping those of different languages apart, the objective encourages word representations to reflect not only the phonetic properties of individual forms, but also the broader phonological profile of the language in which they occur. Enriching the lexical representations with information on the language phoneme inventory, its phonotactics or characteristic syllable structure, arguably adds a deeper phylogenetic signal than what surface form similarity on the concept level would be capable of encoding. The language-level loss can therefore be understood as a form of regularisation that guides the word representation space towards reflecting genealogical structure, rather than phonetic similarity between individual forms alone.

No validation set is used during training due to the model requiring fixed numbers of languages and concepts, complicating the creation of held-out partitions. These are, however, not crucial for training, because the training loss alone is not directly indicative of the quality of the only indirectly related downstream tasks, since the training objective is not fully congruent with those. For the same reason, hyperparameter optimisation was omitted and hyperparameters were chosen either based on the values provided by \citet{wientzek2025b} or determined heuristically (see \autoref{tab:hyperparams}).

The model was evaluated after 15 epochs, each seed running for a total of about 3 minutes, with about 5 more minutes for tree inference. Experiments were run on a single NVIDIA GeForce 3070 Notebook GPU with 8 GB VRAM, AMD Ryzen 9 5900HX CPU, 16 GB RAM and CUDA 13.0. Exact package versions, training scripts, model checkpoints and seeds used for reproducibility can be found in the Supplementary Material \citep{suppmat}.

\begin{table}[htbp]
\centering
\caption{Model hyperparameters.}
\label{tab:hyperparams}
\begin{tabular}{ll}
\toprule
\textbf{Parameter} & \textbf{Value} \\
\midrule
Embedding dimension & 256 \\
Encoder layers & 1 \\
Encoder heads & 4 \\
Learning rate & 0.001 \\
Number of languages & 32 \\
Number of concepts & 30 \\
Temperature & 0.3 \\
Duplication probability & 0.1 \\
Noise probability & 0.5 \\
Input dropout & 0.1 \\
Attention dropout & 0.1 \\
\bottomrule
\end{tabular}
\end{table}

\subsubsection{Evaluation}
\label{sec:eval}

\paragraph{Obtaining language distances}
To compute distances between language pairs, the approach standard in lexical distance-based methods is adopted \citep[see e.g.][]{muller2009,rama2020}. For a language pair $i, j \in L$, the set of concepts $C_{ij}$ for which data are available in both languages is identified. For each concept $k \in C_{ij}$, the cosine distance between the respective word form representations in the two languages is computed, and these distances are averaged across all shared concepts $C_{ij}$ (\autoref{eq:distance}). The word form representations $v_{k}(i)$ and $v_{k}(j)$ are the outputs of the word encoder, as described in \S\ref{sec:model}. In cases where synonyms or alternative transcriptions exist for a given concept in a given language, $v_{k}(l)$ is taken as the mean over the representations of those variants.

\begin{equation}
\label{eq:distance}
dist(i, j) = \frac{1}{N}\sum_{k=1}^{N} d(v_{k}(i), v_{k}(j))
\end{equation}

From the pairwise distances across all language pairs $i, j$, a distance matrix of size $L \times L$ is constructed.

\paragraph{Phylogenetic tree}
BIONJ \citep{gascuel1997}, a clustering method commonly employed in computational biology and based on neighbour-joining, was used to construct trees out of the resulting distance matrix. Following \citet{jager2026}, the quality of the tree is assessed via comparison to the Glottolog tree \citep{glottolog2025}. Tree similarity is quantified with the generalised quartet distance  \citep[GQD]{pompei2011}. This metric takes into account that the Glottolog tree is, in contrast to the ones produced by BIONJ, not fully resolved, i.e., the branches are not strictly bifurcating. As such, the distance is defined as the proportion of differently resolved quartets, i.e., groups of four languages, and total resolved quartets in both trees. As such, a lower GQD value indicates a higher similarity between tree topologies. In addition to the trees inferred via the presented method and the tree by \citet{jager2026}, both a tree based on a pair-Hidden Markov Model \citep[pHMM]{durbin1998} and on normalised Levenshtein Distances (LDN) were used as a baseline for comparison (see \citet{jager2025} for details). For both, the ASJP encodings provided in the data were used for model fitting and distance calculations and the trees were inferred with BIONJ. GQD is calculated with the freely available software QDist \citep[available at \url{https://birc.au.dk/software/qdist/}]{mailund2004}. 

\paragraph{Concept stability}
Diachronic concept stability, i.e., the resistance to lexical replacement over time, has been of central importance in lexicostatistics for decades. Concept lists in databases such as ASJP or Lexibank are designed to consist predominantly of core vocabulary, that is, concepts with high diachronic stability. These Swadesh-style lists \citep{swadesh1955} facilitate the application of the comparative method by reducing the probability that borrowings, semantic shifts, or onomatopoeia introduce noise into the data. Ultimately, more stable word forms allow for deeper looks into the history of language diversification. \citet{rama2020} investigated whether neural word form representations capture this property by computing, for each concept, the standard deviation of the distribution of pairwise distances between its representations across languages as a proxy for stability and comparing the resulting ranking against several concept stability rankings compiled by \citet{dellert2018} (\autoref{tab:concept_stability}). Significant correlations were found for most rankings. The present study follows the same approach, applied to the word representations retrieved from the word encoder (see \S\ref{sec:model}).

Formally, for a concept $k$, the cosine distance $d(v_k(i), v_k(j))$ is computed for all language pairs $i, j \in L$ for which $k \in C_{ij}$, that is, for which the concept is attested in both languages. The stability score $s(k)$ is then defined as the variance of these distances over all such pairs.

The intuition is as follows. For pairs of related languages, stable concepts are expected to yield consistently low distances, since the corresponding word forms are more likely to have been inherited from a common ancestor and to have undergone limited change. Less stable concepts, by contrast, are more likely to have been replaced or altered, and their pairwise distances will therefore be skewed toward high distances with only a small low-distance peak. For unrelated language pairs, distances are expected to be high regardless of concept stability. The combined effect is that the distance distribution of a stable concept is more clearly bimodal, i.e., a larger peak in the low values for related pairs and high values for unrelated ones. This bimodal spread across both ends of the range yields a higher variance than the unimodal distribution, skewed toward high distances without a comparable low-distance mode, expected for less stable concepts. Therefore, a higher variance should indicate greater stability.

Because some concepts occur in only a small number of datasets, their stability scores may be artificially inflated. Datasets tend to cover genealogically related languages within restricted areas, increasing the share of form similarities due to a higher proportion of in-family pairs in the overall sample. To reduce this effect, only concepts attested in at least 25\% of languages in the sample are included in the stability ranking, resulting in 122 ranked concepts.

The rankings used as basis for comparison (see \autoref{tab:concept_stability}) vary along two dimensions that are relevant for interpreting the correlations: whether stability is assessed via cognacy judgements or directly from word form similarity, and how broad the underlying language sample is. \citet{pagel2007}, \citet{peust2013}, and \citet{petroni2011} all rely on cognacy or string similarity within one or a small number of language families, possibly introducing genealogical or areal biases. The three WOLD-based scores from \citet{haspelmath2009} - age, borrowability, and morphological simplicity - and Tadmor's \citep{tadmor2009} composite thereof are expert judgements drawn from a globally diverse but comparatively small language sample. Holman et al. \citet{holman2008} provide automated cognacy-based stability estimates from the ASJP database, covering a wide global sample. \citet{rama2014} also use the full ASJP database and quantify the within-family phonological variation of concepts via n-gram entropy. \citet{dellert2018} estimate morphological basicness from information content on the NorthEuraLex database \citep{dellert2019}, covering languages from 20 families.

\section{Results}
\subsection{Phylogenetic inference}
\begin{figure}[!ht]
\centering
\includegraphics[width=0.8\textwidth]{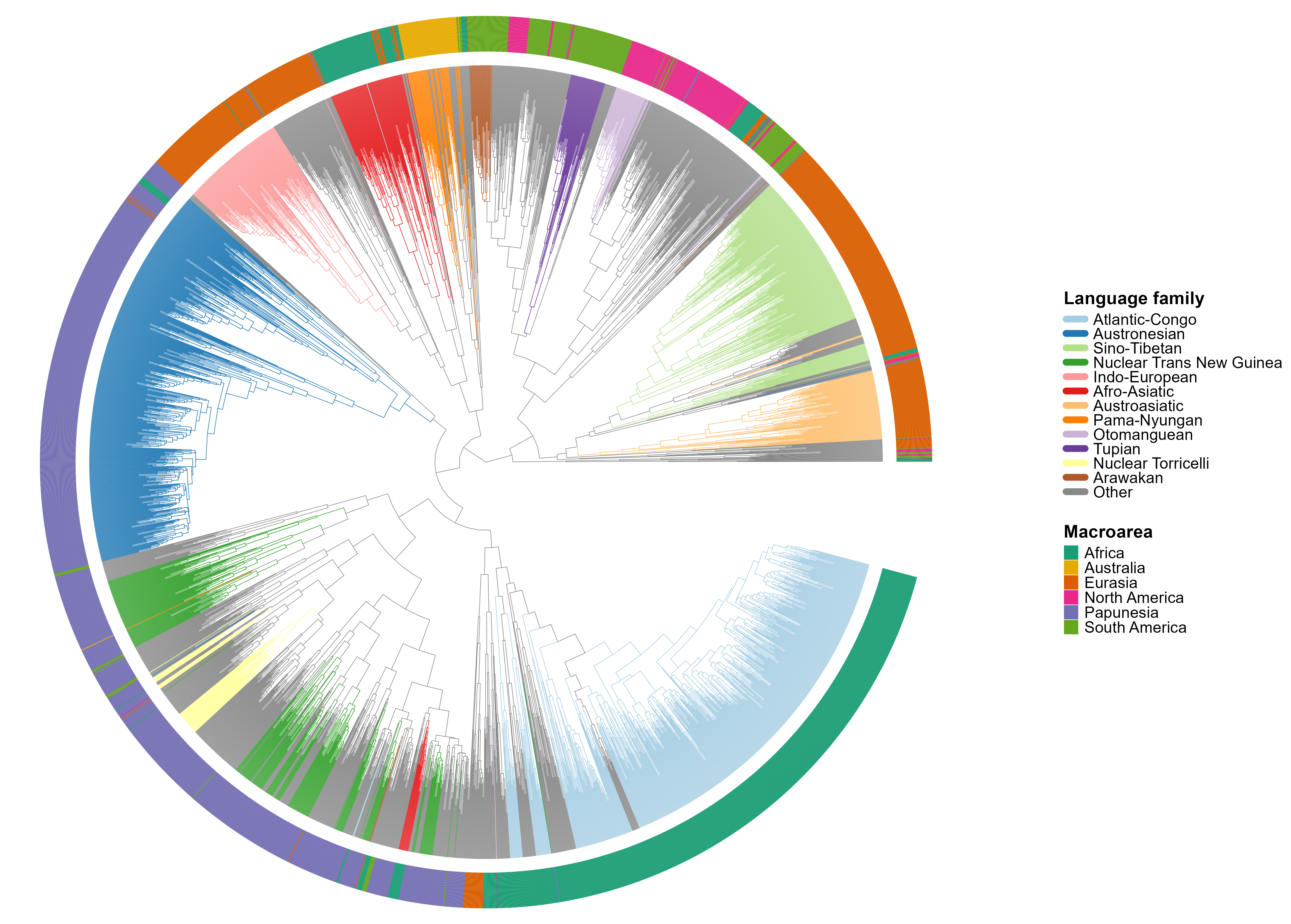}
\caption{Representative phylogenetic tree inferred from the learned lexical representations, rooted with MAD and ultrametricised via penalised likelihood. The tree was visualised with the R package \textit{ggtree} \citep{ggtree}.}
\label{fig:worldtree}
\end{figure}
\autoref{tab:phylogenetic} shows the GQD of the BIONJ tree inferred from the model's distance matrix and the baseline approaches, relative to the Glottolog reference tree. With a mean GQD of 0.0346 (SD = 0.003 across 5 runs), the DualCWE outperforms the MSA approach (0.0433) and the LDN and pHMM baselines. A representative tree, rooted using Minimal Ancestor Deviation \citep[MAD]{tria2017} and ultrametricised via penalised likelihood \citep{sanderson2002} with the settings described in \citet{jager2026}, is shown in \autoref{fig:worldtree}. The resulting MAD score of 0.013, compared to 0.060 for the tree in \citet{jager2026}, indicates greater confidence in the rooting and the molecular clock assumption. Visually inspecting the tree, most major language families generally cluster coherently. Among the large families, Nuclear Trans New Guinea is the most visibly fragmented. This is not surprising, given that the internal subgrouping of this family remains contested in the literature \citep{greenhill2024}, hinting at a weak genealogical signal in the lexical data for the model to pick up. Notably, the mainland Southeast Asian contact zone shows a cleaner separation between Sino-Tibetan and Austroasiatic languages, a primary source of error identified by \citet{jager2026}. 

\begin{table}[htbp]
\centering
\caption{Generalised quartet distance (GQD) of trees inferred from normalised Levenshtein distance (LDN), a pair-Hidden Markov Model (pHMM), multiple sequence alignment (MSA) and lexical representations (DualCWE). LDN, pHMM and DualCWE are distance-based methods with trees inferred via BIONJ, MSA is a character-based method using maximum likelihood inference.}
\label{tab:phylogenetic}
\begin{tabular}{llll}
\toprule
\textbf{Method} & \textbf{GQD} & \textbf{F1} \\
\midrule
LDN & 0.0468 & 0.9639 \\
pHMM & 0.0692 & 0.9636 \\
MSA \citep{jager2026} & 0.0433 & 0.959 \\
DualCWE (own) & \textbf{0.0346} (0.003) & 0.955\\
\bottomrule
\end{tabular}
\end{table}

Following \citet{jager2026}, I additionally report the macro-averaged family F1-score, which for each Glottolog family identifies the clade in the inferred tree maximising the F1-score between the clade's leaf set and the family's member set, then averages the resulting scores equally across families regardless of size. This complements GQD by directly measuring how coherently the tree groups languages into their presumed families. The F1-scores are consistently high across all methods, with LDN performing slightly better than the others. This suggests that family-level groupings are recovered with comparable accuracy across approaches, and that the observed differences in GQD stem from finer-grained topological variation within families, rather than from broad misclassification of languages. There is also a significant positive correlation between pairwise cosine distances and the corresponding great-circle distances between language locations (r = 0.2021, p < 0.001), consistent with the geographic signal observed by Jäger \citep{jager2018} for PMI-based distances (r = 0.193). 

The Supplementary Material \citep{suppmat} contains all results, alongside family-wise GQD values. These were obtained by extracting the relevant submatrix of pairwise distances from the global distance matrix, reinferring a BIONJ tree for each family from the respective submatrix, and comparing it against the Glottolog reference tree pruned to the respective languages.

\subsection{Concept stability}
\label{sec:concept_stability}

\begin{table}[htbp]
\small
\centering
\caption{Spearman rank correlations between the model-derived concept stability ranking (with higher variance resulting in ranks closer to 1) and established stability rankings. Superscript $^{+}$ indicates rankings where higher values correspond to greater stability; $^{-}$ indicates rankings where lower values correspond to greater stability. Column ``Overlap'' shows the number of concepts from the ranking that overlap with the model's 122 ranked concepts (total size of the ranking in parentheses). Significant correlations (p $<$ 0.05) are shown in bold.}
\label{tab:concept_stability}
\begin{tabular}{llll}
\toprule
\textbf{Ranking} & \textbf{Overlap} & \textbf{Spearman r} & \textbf{p-value} \\
\midrule
\cite{holman2008}$^{+}$ & 69 (100) & $\mathbf{-0.4907}$ & $\mathbf{0.0000}$ \\
\cite{pagel2007}$^{-}$ & 97 (200) & $\mathbf{0.3172}$ & $\mathbf{0.0015}$ \\
\cite{petroni2011}$^{+}$ & 61 (100) & $\mathbf{-0.2801}$ & $\mathbf{0.0288}$ \\
\cite{peust2013}$^{-}$ & 101 (185) & $\mathbf{0.3449}$ & $\mathbf{0.0004}$ \\
\cite{rama2014}$^{-}$ & 69 (100) & $\mathbf{0.7204}$ & $\mathbf{0.0000}$ \\
\cite{tadmor2009}$^{+}$ & 59 (100) & $-0.1980$ & $0.1327$ \\
Tadmor (replica)$^{+}$ & 122 (1463) & $0.0729$ & $0.4248$ \\
WOLD age score \citep{haspelmath2009}$^{+}$ & 122 (1463) & $-0.0324$ & $0.7232$ \\
WOLD borrowed score \citep{haspelmath2009}$^{-}$ & 122 (1463) & $-0.1584$ & $0.0814$ \\
WOLD simplicity score \citep{haspelmath2009}$^{+}$ & 122 (1463) & $\mathbf{-0.2412}$ & $\mathbf{0.0074}$ \\
\cite{dellert2018}("inf")$^{-}$ & 122 (769) & $\mathbf{0.4510}$ & $\mathbf{0.0000}$ \\
\bottomrule
\end{tabular}
\end{table}
\autoref{tab:concept_stability} shows the Spearman rank correlations between the word-representation-derived stability ranking and several established rankings. Significant correlations (p < 0.05) are found for 7 of the 11 rankings, with the strongest agreement with \citet{rama2014}, \citet{holman2008}, and \citet{dellert2018} word basicness measure (``inf''). The ranking agrees most strongly with those derived from globally diverse lexical data, and less so with rankings based on single or few families, such as \citet{pagel2007}, \citet{peust2013}, and \citet{petroni2011}. This is expected: the present ranking is itself derived from a globally spanning dataset and is therefore less prone to areal or genealogical biases that can inflate the observable stability of culturally prominent concepts in regionally restricted samples. Rankings derived from WOLD metrics \citep{haspelmath2009}, including \citet{tadmor2009} and its replica, draw on a global language sample but yield weaker or insignificant correlations, possibly reflecting the comparatively small number of languages in WOLD or just the methodological differences.

In general, it is striking that all concepts in the top 20 (\autoref{tab:concept_deviations}) are systematically ranked higher (here and below, higher refers to closer to the top) than they are on average in the comparison rankings - a phenomenon that can conversely be observed for the lower ranks as well. On the other hand, many of these concepts are also predominantly located in the upper part of other rankings, see, for example, I, THOU, THREE, EYE, DIE, or TONGUE, which generally supports the validity of the method.  

\begin{table}[htbp]
\scriptsize
\centering
\caption{Average rank of the 20 highest variance concepts across the 5 seeded runs (``Avg.\ Rank''), with per-ranking deviations from the resulting placement. A negative deviation indicates the concept is ranked higher (more stable) in that specific ranking; a positive deviation indicates it is ranked lower (less stable). The three WOLD scores are omitted for readability and because they are not direct stability rankings; a weighted composition of these scores is captured in the two Tadmor rankings.}
\label{tab:concept_deviations}
\begin{tabularx}{\textwidth}{l*{12}{>{\centering\arraybackslash}X}}
\toprule
\textbf{Concept} & \textbf{Avg. Rank} & \textbf{Holman} & \textbf{Pagel} & \textbf{P \& S} & \textbf{Peust} & \textbf{Rama} & \textbf{inf} & \textbf{Tadmor} & \textbf{Tad. (rep.)} & \textbf{Mean Disagr.} \\
\midrule
FATHER & 1.4 & - & 56 & 26 & 34 & - & 27 & - & 105 & 49.6 \\
MOTHER & 1.6 & - & 59 & 1 & 48 & - & 20 & & 84 & 42.4 \\
I & 3.4 & 2 & -2 & - & 1 & -2 & 1 & 10 & 6 & 2.29 \\
AND & 3.6 & & 79 & - & - & - & -3 & - & 65 & 47.0 \\
HORSE & 6.2 & - & - & - & - & - & 64 & - & 117 & 90.5 \\
FIVE & 6.2 & - & -4 & 20 & -4 & - & 50 & - & 96 & 31.6 \\
THOU & 7.0 & 7 & 5 & - & -3 & 10 & -5 & 3 & 22 & 5.57 \\
THREE & 7.2 & - & -7 & -6 & -4 & - & 22 & - & 68 & 14.6 \\
EYE & 12.0 & -3 & 13 & & 19 & -6 & 10 & 44 & 40 & 16.71 \\
DRINK & 12.6 & 22 & 24 & 43 & 40 & -6 & 13 & 22 & 35 & 24.12 \\
NOT & 12.8 & 46 & 6 & -7 & 55 & 25 & -5 & 30 & -9 & 17.62 \\
ARM & 12.8 & - & - & - & - & - & 35 & - & 28 & 31.5 \\
THIS & 13.6 & 45 & 41 & - & 37 & -3 & -6 & 17 & 14 & 20.71 \\
MEAT & 14.0 & 49 & 24 & 9 & 68 & 37 & 45 & 2 & 18 & 31.5 \\
DIE & 14.2 & -11 & 2 & 1 & -7 & -10 & 75 & - & 101 & 21.57 \\
SEVEN & 14.8 & - & - & - & -1 & - & 94 & - & 94 & 62.33 \\
BRANCH & 15.4 & - & - & - & - & - & 32 & - & 46 & 39.0 \\
WING & 18.4 & - & 50 & 41 & 48 & - & 82 & -3 & 10 & 38.0 \\
HERE & 19.8 & - & 63 & - & - & - & 45 & - & 83 & 63.67 \\
TONGUE & 20.2 & -4 & -10 & 18 & -12 & 23 & 65 & -14 & -4 & 7.75 \\
\bottomrule
\end{tabularx}
\end{table}

\begin{figure}[!ht]
\centering
\includegraphics[width=0.8\textwidth]{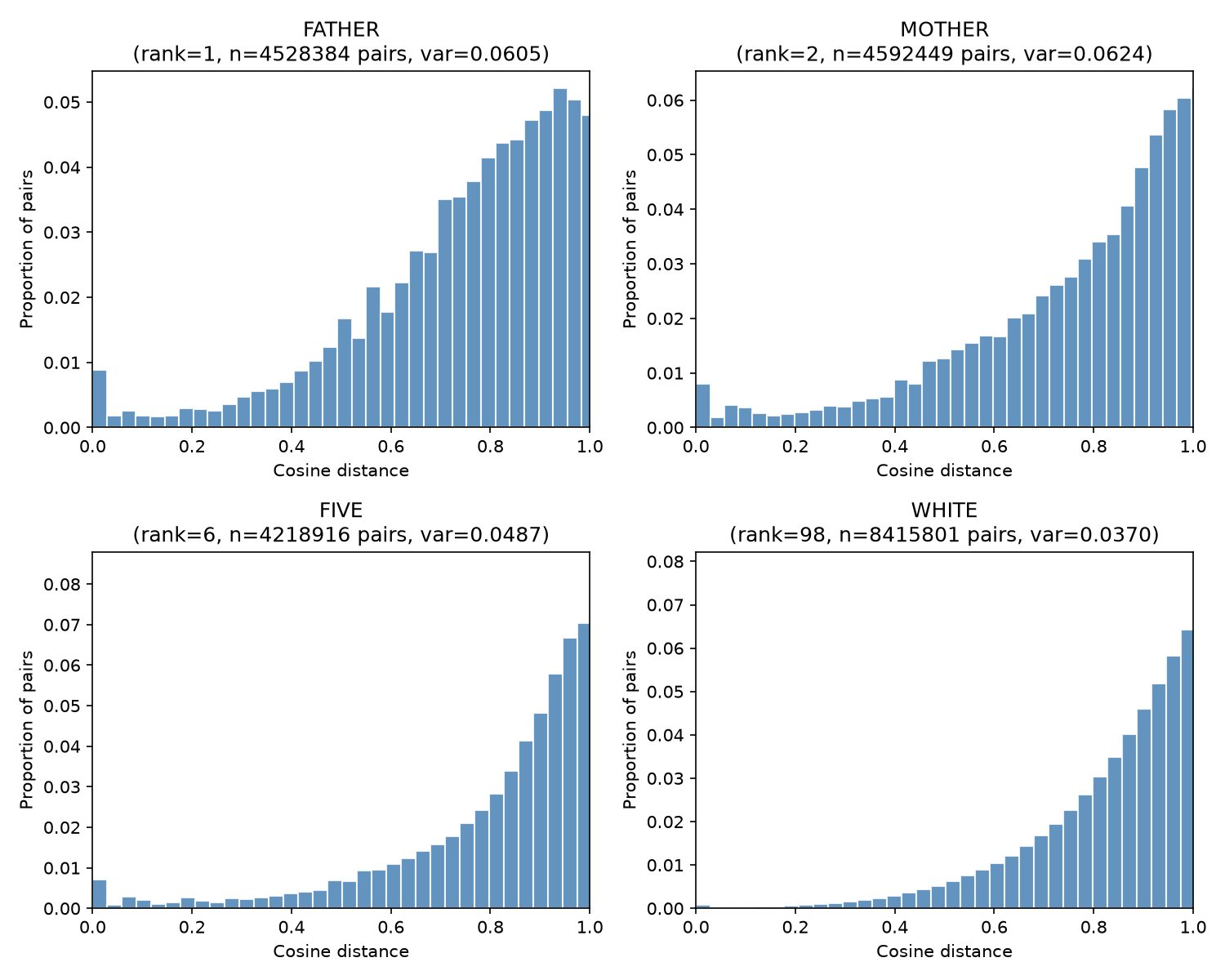}
\caption{Distribution of cosine distances for representations of concepts of different ranks from a single model run. The ranks are based on the average rank across seeds.}
\label{fig:concdist}
\end{figure}

Some concept placements are presumably ranked too highly due to factors that do not directly reflect diachronic stability. For example, both FATHER and MOTHER consistently occupy the first two positions. A plausible explanation is that these concepts are expressed by forms that are variants of the early babble sequences \textit{/mama/} and \textit{/papa/} in a wide range of languages. This results in highly similar forms across many language pairs, including pairs of languages that are not genealogically related, whereas languages that employ conventional lexical forms (such as \textit{mother} or \textit{father}) exhibit the usual pattern of similarity distribution, with high sequence similarity being primarily associated with genealogical relatedness. Note that this does not take into account the - albeit most likely negligible - potential effects of sound symbolism, which has been attested for these two concepts \citep{johansson2020}. Consequently, compared to concepts ranked lower (e.g., WHITE), the distance distributions for FATHER and MOTHER show a higher density throughout the lower half of the range and a pronounced peak at very low distances, resulting in increased variance (\autoref{fig:concdist}). The methodology applied here does not include a mechanism for filtering out factors such as non-diachronic or simply chance similarities, or sound symbolism. In principle, it can nevertheless be assumed that MOTHER and FATHER are among the most stable concepts - even if they should presumably not be ranked quite as highly as suggested here. 

The very high ranking of HORSE is more surprising, as this is a concept that is not even included in most rankings, presumably because it does not meet the conditions for being a core vocabulary item. In some regions of the world (particularly the Americas and Australia), horses played no role or were simply unknown prior to European colonisation. One possible explanation is that it tends to be a non-analysable word, as suggested by its high WOLD simplicity score - being, in fact, the animal with the highest simplicity score in WOLD. On the other hand, the word forms are not unusually short, but rather roughly average in length (HORSE: 4.61, average: 4.65), so there is no reason to assume an above-average degree of chance similarity. One possible conclusion, therefore, is simply that HORSE is highly stable in languages spoken by cultures that have been domesticating horses for a very long time. In cultures where this is not the case, borrowings from the few relevant colonial languages (e.g., mainly Spanish in the case of South America) are likely to result in high similarity.  

AND, which does rank first in the inf-ranking, but much lower or not at all in others, is also unusually highly placed here. A straightforward factor contributing to its very high ranking could be that it is the shortest concept on average in the dataset (AND: 2.8), making it the most susceptible to chance similarity.  

For other concepts such as ARM, BRANCH, SEVEN, or HERE, their high rankings are less straightforward to explain. These concepts are likely to be attested in most cultures and belong to semantic categories commonly exhibiting higher stability. However, there is no obvious factor that would account for their particularly high placement in the present ranking, requiring further investigation before any conclusions about their diachronic stability can be drawn.

\subsection{Ablation}
\label{sec:ablation}

To assess the contributions of the language encoder and its language embeddings, the feature vector character encoding and the sample size to phylogenetic inference quality, the model was run under additional configurations: without the language encoder and with the hyperparameters of the full model, without the language encoder at the original settings of \citet{wientzek2025b} (4 epochs, temperature of 0.05), without adding a language embedding to the encoded word forms in the language encoder, with the full model using ASJP and IPA character transcriptions, and with the reduced concept sample used in \citet{jager2026}. It is noteworthy that a few languages only had less than 30 concepts from the reduced set. In these cases, their concepts were resampled to guarantee correct batch generation. Since all ablation results fall outside the confidence interval of the main model, a single run per configuration was sufficient to establish a meaningful difference. As shown in \autoref{tab:ablation}, both the language encoder in its current design, the use of the full sample, and the phonetic feature vector encoding contribute significantly to phylogenetic accuracy.

\begin{table}[htbp]
\centering
\caption{Ablation results showing GQD for different model configurations.}
\label{tab:ablation}
\begin{tabular}{lc}
\toprule
\textbf{Setting} & \textbf{GQD} \\
\midrule
Original & \textbf{0.0346} \\
No language encoder & 0.044 \\
No language encoder, 4 epochs, 0.05 temp. & 0.071 \\
No language embeddings & 0.054 \\
ASJP encoding & 0.053 \\
IPA encoding & 0.049 \\
105 concepts & 0.052 \\
\bottomrule
\end{tabular}
\end{table}

\section{Discussion}
This work presented a self-supervised framework for learning word form representations from raw multilingual lexical data. The model's objective is to organise phonetically and phonotactically similar word forms into a coherent representational space. Phylogenetic signal is not directly optimised for, but is a by-product of the model learning to capture the kind of phonological regularity that, in historically related languages, reflects shared inheritance. This is achieved without any access to cognacy judgements, family labels, geographic information, or word alignments. That such signal emerges robustly enough to support global-scale phylogenetic inference competitive with existing automated methods is therefore the central finding of this work and underlines the potential of distance-based approaches for fast tree inference and hypothesis testing.

The contrastive training objective does not guarantee that the learned distances reflect genealogical relatedness rather than areal similarity and the significant correlation between cosine distances and great-circle distances indicates that some geographic signal remains in the representations. Given that the overwhelming majority of language pairs in a global sample are too distant to have had meaningful contact, this correlation is driven by geographically proximate pairs, for which it is relatively pronounced. The model, receiving no information about the geographic or genealogical relations between languages, has no mechanism for separating inherited phonological similarity from contact-induced convergence. Disentangling these two components is a natural direction for future work, for instance through contrastive losses designed to separate genealogically unrelated but geographically proximate languages or through integrating a contact detection component that can filter areal signals before distance computation. It is also worth mentioning that branch lengths in the inferred trees reflect relative divergence in the learned representational space and have no directly interpretable unit. This is a general property of distance-based phylogenetic methods without external calibration points and is not specific to the present approach. However, it has no relevance to the quality of the topology.

Using GQD to compare trees of this scale comes with its own shortcomings. While a GQD of 0.035 can be read as "96.5\% of quartets in the inferred tree are consistent with the resolved quartets in the Glottolog reference tree", it is worth noting that an overwhelming majority of quartets in the tree contain languages from at least two families. As such, even a GQD as low as the one reported here is not very informative about the family-internal structure, which is represented by comparably few quartets. In fact, the average family-wise GQD for families consisting of more than 50 languages is about 0.12 \citep[see Supplementary Material;][]{suppmat} and higher on average for smaller families. While not inaccurate, a model tuned on family-specific data would almost certainly be capable of achieving a better resolution. Nonetheless, GQD remains a valuable metric to assess the overall tree quality and ensure comparability.

The model produces both word and language representations, but evaluation is restricted to the word-level output. The language encoder's per-language embedding vectors are best understood as identifiers, allowing the training objective to distinguish two views of the same language from views of different languages, but are not required to place genealogically related languages closer together than unrelated ones. Genealogical signal resides in the lexical items themselves rather than in any language-specific parameter. Evaluating language-level representations, at least in this specific model architecture, would therefore not reflect what the model has learned about the relationship between phonological form and historical relatedness.

The model represents IPA characters as 39-dimensional phonetic feature vectors rather than one-hot encodings or learned character embeddings, and the ablation results confirm that this choice is strongly beneficial. Feature vectors place characters in a structured, linguistically meaningful space, where phonetically similar sounds are geometrically proximate and the dimensions of the space correspond properties such as place and manner of articulation. In data-sparse settings and with Transformer architectures that require lots of data to learn meaningful internal structure, providing this prior organisation from the start is especially valuable. The model can immediately attend to phonological similarity rather than first having to recover it from a discrete input. The same advantage has been observed in related work on lexical data. \citet{wientzek2025a} found consistent improvements from feature vector encoding over alternative character representations in phonological reconstruction and reflex prediction, and \citet{wientzek2025b} found the same in the borrowing detection setting.

The decision to use all 210 pre-selected concepts rather than the 105 used by \citet{jager2026} proves beneficial, as confirmed by the ablation results. A larger concept set yields higher pairwise overlap between languages, providing more evidence per distance estimate. As shown, the model also acquires a notion of concept stability and more stable concepts may contribute more consistently to pairwise distances, even without explicit weighting. The optimal concept count for this approach may differ from what is optimal for maximum likelihood inference on a character matrix and exploring this further is a straightforward extension. Using variance as a proxy for concept stability, however, comes with the same shortcomings that other methods of automated derivation of diachronic concept stability bring with them. Namely, shorter word forms (e.g., AND) or word forms shaped by universal tendencies (e.g., MOTHER and FATHER) will naturally lead to false cognates and to variance overestimating the stability. Still, the rank correlation of the ranking presented here and established rankings suggests that variance in learned lexical representations can be a good approximation to concept stability.

Hyperparameters were chosen based on training dynamics and prior experience rather than systematic optimisation. The batch composition in particular, 32 languages at 30 concepts each, was fixed early in development and not varied. This leads to a quite obvious imbalance when comparing model inputs for languages with the minimum of 40 total concepts and languages with over 200. For the former, the language encoder sees a majority of available concepts at each epoch while for the latter, the concept sets in two batches may be completely different. The concrete effects of this remain to be explored. So while the chosen settings yield strong results, they are unlikely to be optimal, and targeted tuning may yield further improvements. The model was also designed and evaluated at global scale, using a large, typologically diverse sample of around 3400 languages. Family-wise performance, shown in the Supplementary Material \citep{suppmat}, is acceptable, but applying the model to a single family of 50 languages would require revisiting the batch composition and related parameters, as the distributional conditions differ from the global setting.

The most practically significant aspect of the approach is its accessibility. The full inference pipeline runs in minutes on a standard notebook GPU, compared to over 100 hours on a 144-core server for the MSA-based workflow of \citet{jager2026}. Also, requiring only raw IPA transcriptions removes cognacy annotation as a prerequisite entirely. As noted in the introduction, cognacy judgement is slow, expertise-intensive, and susceptible to expert disagreement in ways that complicate replication. Methods that automate or bypass this step can be applied to a much wider range of language groups, including all for which merely a word list exists. A reliable and accessible phylogeny can also be helpful for applying the phylogenetic comparative method to typological and historical questions that would otherwise require trees established by experts \citep{jager2018}. Beyond phylogenetic inference, the lexical representations and language distances derived from them may potentially prove useful for a range of related tasks, such as the initially mentioned cognate discovery, borrowing and contact detection, language clustering and dialectometry, or transfer language selection for low-resource NLP. The present results, together with the borrowing detection results in \citet{wientzek2025b}, suggest that the learned embedding space may capture historically relevant patterns beyond surface similarity. Whether the representations generalise well to these tasks without further adaptation is an open question that future work should address.

\section{Conclusion}
In summary, this paper presents a neural network that learns representations of lexical data, carrying information about the diachronic stability of concepts and the relationships between languages. For phylogenetic inference, the main advantages of the approach are its computational efficiency and largely reduced runtime compared with previously proposed methods, while also achieving improved topological quality. The model presented here is intended primarily as a proof of concept for the potential of neural networks for word and language representation and leaves room for further improvement. More broadly, the learned representations may be useful for a range of applications in historical linguistics, NLP, and at the interface between the two, the potential of which remains to be explored.

\section*{Data availability}
Zenodo: Self-Supervised Lexical Representation Learning for Fast, Large-Scale Phylogenetic Inference: Supplementary Material \citep{suppmat}. The supplementary data is also available at \url{https://github.com/TGH-2020/DualCWE}.

\noindent This project contains the following underlying data and code:
\begin{itemize}
  \item data: underlying linguistic data (word lists, metadata, stability rankings)
  \item src: source code for the project
  \item checkpoints: saved model checkpoints
  \item out: intermediate output files generated during processing
  \item results: final results of the project
  \item run\_experiment.py: script to run the experiments and reproduce the results
\end{itemize}
Data are available under the terms of the Creative Commons Attribution 4.0 International license (CC-BY 4.0).

\section*{Acknowledgments}
This research was supported by the DFG Center for Advanced Studies in the
Humanities Words, Bones, Genes, Tools (DFG-KFG 2237).  
The author wishes to thank Gerhard Jäger for providing the trees based on pHMM and Levenshtein Distances and, especially, his helpful comments on the first draft. DeepSeek V4 Flash was used to assist with code readability, light language editing, and formatting of the manuscript.

\bibliography{references}

\end{document}